\documentclass[a4paper,fleqn]{cas-sc}

\usepackage[numbers]{natbib}

\usepackage{soul, color}
\usepackage[T1]{fontenc}
\usepackage[utf8]{inputenc}

\usepackage{amsmath,amsfonts,amssymb}
\DeclareMathAlphabet{\mathbbold}{U}{bbold}{m}{n}

\usepackage{bbm}

\usepackage{xcolor}

\usepackage{colortbl}
\definecolor{winstrong}{rgb}{0.9,1,0.9}  

\definecolor{darkgreen}{RGB}{0, 90, 0} 
\newcommand{\ccheck}{\textcolor{darkgreen}{\checkmark}}

\newcommand{\phantommark}{\hphantom{\,\ccheck}}
\usepackage{pifont} 

\usepackage{prettyref}
\newrefformat{fig}{Figure~\ref{#1}}
\newrefformat{tab}{Table~\ref{#1}}

\def\tsc#1{\csdef{#1}{\textsc{\lowercase{#1}}\xspace}}
\tsc{WGM}
\tsc{QE}
\tsc{EP}
\tsc{PMS}
\tsc{BEC}
\tsc{DE}

\begin{document}
\let\WriteBookmarks\relax
\def\floatpagepagefraction{1}
\def\textpagefraction{.001}
\shorttitle{Bridging the Modality Gap in Forensic Image Retrieval}

\title [mode = title]{Bridging the Modality Gap in Forensic Image Retrieval}                      

%

\author[1]{Ricardo González-Gazapo}
\fnmark[1]
\ead{rgazapo@cenatav.co.cu}

\author[1]{Annette Morales-González}[orcid=0000-0003-2716-3144]
\cormark[1]
\ead{amorales@cenatav.co.cu}
\cortext[cor1]{Corresponding author}

\author[1]{Yoanna Martínez-Díaz}
\ead{ymartinez@cenatav.co.cu}

\author[1]{Heydi Méndez-Vázquez}
\ead{hmendez@cenatav.co.cu}

\author[2]{Milton García-Borroto}
\ead{milton.garcia@fisica.uh.cu}

\affiliation[1]{organization={Advanced Technologies Application Center (CENATAV)},
	city={Havana},
	country={Cuba}}
\affiliation[2]{organization={Centro de Sistemas Complejos, Facultad de Física, Universidad de La Habana},
	city={Havana},
	country={Cuba}}
%
%

%

\begin{abstract}
Automated image retrieval plays an increasingly critical role in modern forensic analysis, supporting investigative workflows that rely on efficient comparison of visual evidence. While prior work has focused primarily on developing and optimizing multimodal retrieval systems, limited attention has been paid to evaluating the forensic applicability of these technologies across diverse real-world scenarios. In this study, we present a unified retrieval framework adapted to four key forensic tasks: (1) tattoo image retrieval given a tattoo query image; (2) tattoo retrieval guided by human-expert textual descriptions, modelling the common situation where a witness verbally describes a tattoo; (3) tattoo retrieval from hand-drawn sketches; and (4) face retrieval from forensic face sketches. Our system leverages a multimodal large language model (MLLM) to automatically generate structured textual descriptions for all queries and gallery images, followed by sentence-transformer embedding for text-based comparison. We evaluate retrieval using visual-only embeddings, text-only embeddings and a multimodal fusion strategy that combines text- and image-based similarity scores derived from state-of-the-art visual feature extractors relevant to each task. The fusion of modalities consistently improves retrieval precision and robustness, especially in scenarios where visual information is limited or noisy (e.g., sketches, partial tattoos, or fragmented witness statements).
This work highlights the forensic value of a unified multimodal retrieval pipeline and demonstrates how modern MLLMs can operationalize challenging forensic tasks that traditionally rely on manual expert analysis. Our results position multimodal retrieval as a promising tool for supporting investigative workflows involving tattoos, facial composites, and witness descriptions.
\end{abstract}

%


\maketitle

\section{Introduction}

Forensic investigations frequently require experts to compare visual evidence originating from heterogeneous sources—photographs, witness-provided descriptions, composite sketches, and partial or degraded imagery. Tattoos and faces are among the most distinctive visual biomarkers used in criminal identification, immigration control, and victim recognition~\cite{CriminalisticTattoo2020,TattooCrimeSolving2021}. However, the variability in image quality, lighting conditions, drawing styles, and human memory recall poses significant challenges for reliable manual comparison.

In practice, forensic experts often confront several recurrent scenarios: a surveillance camera captures a tattoo only partially; a witness provides a natural-language description of a subject’s tattoo; a hand-drawn sketch is created to approximate a tattoo or a suspect’s face; or investigators possess a query image and must search large databases for matches. Each of these situations demands a retrieval system capable of interpreting heterogeneous inputs while remaining robust to noise, ambiguity, and stylistic variation~\cite{ForensicTattoo2024}. Traditional image-only retrieval systems are insufficient for handling the wide range of representations inherent to real forensic evidence \cite{Janier_TattTRN2024,Miguel_CIARP2019}.

Recent advances in multimodal large language models (MLLMs) have opened new opportunities for bridging these heterogeneous information sources~\cite{CIRSurvey2025}. While the integration of LLMs is already gaining traction in broader digital forensics for automating textual and log analysis~\cite{LLM_DF_Survey2025}, their potential for unifying visual forensic evidence remains largely unexplored. MLLMs are capable of producing structured semantic descriptions of images, capturing high-level content that complements low-level pixel-based features. By transforming images, sketches, and natural-language descriptions into a unified textual space, they offer a flexible foundation for forensic-oriented search. Meanwhile, sentence-transformer models provide a highly discriminative embedding space for comparing text across diverse styles and modalities.

In previous works \cite{MLLMTattoo_CIARP2025,MultimodalTattoo_IWAIPR2025}, authors have developed multimodal retrieval pipelines based on automatic textual description and embedding-based comparison. In this work, we shift the focus from algorithmic development to real-world forensic applicability. Specifically, we conduct an assessment of how  such a unified retrieval framework performs across four major forensic scenarios:

\begin{enumerate}
	\itemsep=0pt
	\item Tattoo-to-tattoo retrieval: searching a tattoo database using a direct tattoo image.
	\item Human description-to-tattoo retrieval: retrieving tattoos based on expert-formulated witness descriptions.
	\item Sketch-to-tattoo retrieval: matching hand-drawn tattoo sketches to photographic tattoo collections.
	\item Sketch-to-face retrieval: identifying possible matches for forensic face sketches.
\end{enumerate}  

All retrieval tasks share the same underlying principle: each image (query and gallery) is described by an MLLM, encoded via a sentence transformer, and compared through textual similarity. When visual embeddings are also available, we combine textual and visual similarity scores, taking advantage of their complementary strengths. This unified structure allows the pipeline to seamlessly adapt to a wide range of forensic evidence formats without requiring task-specific architectures, underscoring the importance of the zero-shot performance of MLLMs in enabling robust retrieval without the need for domain-specific training data.

This paper contributes to forensic science literature in several ways:

\begin{itemize} 
	\item First, it demonstrates the practical value of multimodal semantic description for forensic retrieval tasks where visual evidence is incomplete, ambiguous, or stylized—particularly sketches and witness descriptions. 
	\item Second, it shows that text-only embeddings already provide strong performance when visual features are unreliable or unavailable, supporting investigative workflows that depend solely on verbal information 
	\item Third, it illustrates how combining textual and visual embeddings enhances accuracy across all tasks, providing a robust multi-evidence framework aligned with forensic best practices.
	\item Finally, it provides a single methodology capable of handling multiple forensic modalities under one umbrella, reducing fragmentation and simplifying operational deployment. 
\end{itemize}

By centering the discussion on operational relevance rather than model architecture, this work aims to provide both forensic practitioners and technology developers with new evidence about the emerging potential of language-aware visual search in criminal investigations. This focus responds to calls for AI systems in forensics that are not only accurate but also transparent and aligned with investigative workflows~\cite{AIAlignmentMobileForensics2024}.

\section{Related work}
This section reviews prior research relevant to the proposed forensic image retrieval framework, emphasizing studies on tattoo and face sketch identification, multimodal retrieval strategies, and the emerging use of multimodal large language models (MLLMs) in security contexts. The discussion focuses on works aligned with operational forensic scenarios where queries may be incomplete, descriptive, or non-photographic.

\subsection{Tattoo retrieval in forensics}

Tattoo recognition has evolved from early methods based on handcrafted visual features, such as SIFT descriptors and shape-based contour matching, to modern deep learning approaches \cite{TattooSIFT15,Miguel_CIARP2019,xu2021tattoos}. With the rise of convolutional neural networks (CNNs), models like ResNet and MobileNet have become standard for extracting discriminative representations \cite{Miguel_MTAP2022,Yolov5ResnetTattoo2024}. These approaches generally leverage off-the-shelf pre-trained CNNs, sometimes enhanced with spatial weighting \cite{Miguel_MTAP2022}, or adapt networks through transfer learning on tattoo-specific data \cite{TattooClassif2022}. Notable advances include a Faster R-CNN-based framework for joint detection and representation \cite{JointDetRecogWebTattoo2019}, and the Tattoo Template Reconstruction Network (TattTRN), which maps inputs to canonical templates \cite{Janier_TattTRN2024}. However, a major bottleneck remains the scarcity of large, publicly available datasets, limiting reproducibility \cite{Janier_TattTRN2024,Miguel_MTAP2022}.

Despite these advances, a critical limitation persists: current methods operate primarily on low- or mid-level visual features without capturing semantic content. Tattoos often encode rich symbolic meaning (e.g., religious icons, gang affiliations) and combine heterogeneous artistic styles \cite{TattooHolisticUnderstanding2022}. These elements are crucial for forensic interpretation, providing investigative leads beyond mere visual similarity \cite{TatuajesForense_2025}. Existing CNN-based approaches treat tattoos as generic visual patterns, failing to reason about high-level attributes. Consequently, they struggle when visual appearance varies but semantic meaning remains consistent—precisely the conditions encountered when matching witness descriptions or partial imagery. This semantic gap motivates the adoption of language-mediated representations.

\subsection{Forensic sketch recognition}

Matching forensic face sketches to photographic galleries is a classical heterogeneous recognition problem. Early approaches relied on handcrafted visual descriptors (LBP, HOG, SIFT) to extract modality-invariant features \cite{FaceSketch2012}. Deep learning subsequently improved performance via joint embedding models learning shared representations between sketches and photographs \cite{Vazquez2019LocalDF}. However, forensic face sketch recognition remains challenging due to sketching style variability and limited training datasets. Most forensic datasets contain only a single sketch per subject, increasing overfitting risks.

To mitigate the modality gap, generative approaches using GANs transform sketches into photo-like images, allowing traditional recognition systems to operate on synthesized photographs. Recent research also explores generating facial images directly from textual descriptions. Parallel challenges arise in tattoo identification when sketches are used instead of photographs. The WebTattoo dataset \cite{JointDetRecogWebTattoo2019} provides a resource for this, including paired sketches and images. Early work applied SIFT-based matching to this setting \cite{Han2013TattooBI}, while recent research has explored deep cross-modal approaches using Siamese architectures \cite{SketchCrossModalTattoo2021}. However, these supervised approaches require extensive training, representing a significant bottleneck given the scarcity of specialized forensic datasets required for effective learning. While vision–language models have recently aligned facial representations with semantic descriptions, previous work has not explored the joint use of MLLM-generated textual descriptions with visual representations to improve sketch-based recognition. This gap motivates the multimodal framework proposed here.

\subsection{Composed multimodal retrieval approaches}

While the visual and sketch-based methods reviewed above address specific modality gaps, they remain largely restricted to perceptual similarity derived from appearance. This limits their ability to capture higher-level semantic characteristics routinely considered by forensic practitioners, such as symbolic meaning or thematic content. To address this, research in general multimedia analysis has moved beyond simple cross-modal translation \cite{CrossModalSurvey2024} toward Composed Image Retrieval (CIR), which enables retrieval by combining a reference image with accompanying text \cite{CIRSurvey2025}. Unlike standard cross-modal tasks that replace one modality with another, CIR allows for the joint consideration of complementary evidence—a capability highly relevant to forensic workflows. However, existing CIR approaches typically divide into supervised methods relying on annotated triplets, and zero-shot methods operating without labeled data. In forensic tattoo analysis, supervised approaches are difficult to apply due to data scarcity and ethical constraints \cite{Janier_TattTRN2024,Miguel_MTAP2022}, making zero-shot CIR methods particularly relevant.

Recent zero-shot approaches rely on pre-trained vision–language models. For example, CIReVL \cite{CIReVL2024} generates textual descriptions and recomposes them to approximate target semantics, while LDRE \cite{CIR_LDRE2024} employs dense image descriptions and semantic aggregation. These methods demonstrate that textual representations can encode information not reliably captured by visual features. A related training-free framework combines visual and textual representations using weighted fusion \cite{ZSCIR_WeightedFusion2024}. While this enhances semantic coverage, it increases computational cost. In contrast, our approach is motivated by forensic retrieval, where multimodal techniques remain unevaluated. We integrate automatically generated textual descriptions with visual representations, adopting a multiplicative similarity strategy that reinforces agreement between modalities rather than weighted averaging. Furthermore, we generate a single textual description per image, balancing semantic expressiveness and computational efficiency for operational workflows.

\subsection{Multimodal LLMs for forensic or security applications}

Multimodal Large Language Models (MLLMs) are gaining attention in forensic science due to their capacity to integrate heterogeneous evidence. Unlike traditional tools specializing in single data types, MLLMs process images and language in a unified, semantically rich framework, enabling context-aware reasoning that mirrors human expert workflows. Complementary research in digital forensics highlights the broader potential of LLMs to alleviate investigative workloads through automation in code generation, timeline reconstruction, and log analysis, though challenges regarding bias, explainability, and legal admissibility remain \cite{LLM_DF_Survey2025}. Emerging multimedia applications include image forgery detection and deepfake analysis \cite{VLForgery2025}, as well as training-free pipelines like the In-Context Forensic Chain that integrate multimodal reasoning without task-specific training \cite{chen2025trainingfreeincontextForensic}.

Benchmarking efforts highlight both the promise and limitations of MLLMs in forensic science. Evaluations of state-of-the-art models on structured forensic questions reveal that while multimodal reasoning enhances domain understanding, challenges remain in visual interpretation and nuanced inference \cite{BenchmarkingMLLMsForensic2025}. Complementing this, recent work on AI alignment in mobile forensics underscores the necessity of robustness and governance in image categorization tools to ensure legal admissibility \cite{AIAlignmentMobileForensics2024}. While LLM adoption is growing in digital forensics (textual/log evidence), their application to \textit{visual} forensic retrieval remains largely unexplored. Retrieval tasks involving tattoos and facial sketches frequently rely on incomplete or non-photographic queries not well addressed by purely visual similarity. This work responds to this gap by framing multimodal language–vision modeling within realistic forensic retrieval scenarios, showing how semantically rich textual descriptions complement established visual features.

\section{Forensic retrieval scenarios}

Forensic image retrieval tasks differ fundamentally from conventional multimedia retrieval problems due to the nature of the available evidence, the variability in query quality, and the need for interpretable and reproducible results. In many investigative contexts, queries are not well-defined photographic images but rather partial, degraded, or indirect representations derived from human observation, memory, or secondary documentation. This section describes the forensic scenarios addressed in this work and clarifies how each reflects realistic investigative workflows encountered in law enforcement and forensic casework.

\subsection{Scenario 1: Tattoo image retrieval from a photographic query }
Tattoo image retrieval using a photographic query represents the most direct and traditionally studied forensic scenario. In this setting, a tattoo image is acquired from a suspect, victim, or surveillance source and used to search a database of known tattoo images. Despite being image-based, this scenario remains challenging due to variations in illumination, skin deformation, aging, occlusions, camera viewpoint, and partial visibility. Tattoos may be captured under non-cooperative conditions, often using mobile devices, resulting in inconsistent quality.

From a forensic perspective, even when a photographic query is available, investigators frequently rely on semantic attributes—such as motifs, symbols, textual elements, or stylistic patterns—to guide their search and interpretation \cite{TatuajesForense_2025}. The proposed framework reflects this practice by augmenting visual similarity with automatically generated textual descriptions, enabling retrieval that aligns more closely with expert reasoning rather than purely low-level appearance matching.

\subsection{Scenario 2: Tattoo image retrieval from expert textual descriptions }
In many forensic investigations, no tattoo image is available at the time of inquiry. Instead, information is provided through textual descriptions produced by witnesses, victims, or forensic experts, often based on memory or partial observation. These descriptions may include references to symbolic content (e.g., religious imagery, animals, letters), approximate location on the body, stylistic elements (e.g., tribal, minimalist), or perceived meaning.

This scenario is particularly relevant in missing person cases, victim identification, and early-stage criminal investigations, where textual records constitute the primary source of information \cite{TatuajesForense_2025}. Traditional tattoo recognition systems are not designed to operate on text-based queries, limiting their applicability in such contexts. By encoding expert descriptions into a shared embedding space, the proposed approach enables retrieval that is directly compatible with the narrative-based evidence commonly encountered in forensic reports.

\subsection{Scenario 3: Tattoo image retrieval from sketch-based queries  }
Tattoo sketches are frequently used in forensic practice when witnesses or victims attempt to visually reconstruct a tattoo from memory, often with the assistance of an investigator or forensic artist. These sketches typically capture high-level structural or semantic aspects of the tattoo while lacking fine-grained visual details such as texture, color gradients, or precise proportions.

Sketch-based tattoo retrieval presents substantial challenges due to the large domain gap between hand-drawn representations and photographic images. From a forensic standpoint, sketches are inherently subjective and may emphasize salient symbolic features over exact appearance. The framework proposed in this work addresses this gap by relying on semantic descriptions derived from both sketches and gallery images, enabling retrieval driven by shared conceptual content rather than strict visual correspondence.

\subsection{Scenario 4: Face image retrieval from facial sketches }
Facial sketch-based retrieval is a well-established forensic task, commonly used when photographic evidence of a suspect is unavailable. Sketches may be produced by trained forensic artists or generated using software tools based on witness input. As with tattoo sketches, facial sketches vary widely in accuracy and detail, depending on the witness’s recollection and the circumstances of observation.

In operational contexts, investigators often interpret facial sketches semantically, focusing on descriptive attributes such as face shape, prominent features, or distinctive marks. The retrieval scenario considered in this work aligns with this practice by leveraging multimodal descriptions to bridge the gap between sketch representations and photographic face images. Although face sketch retrieval has been extensively studied, its integration into a unified multimodal retrieval framework alongside tattoo-based scenarios reflects a broader forensic objective: enabling consistent retrieval mechanisms across heterogeneous evidence types.

\subsection{Unified perspective on forensic retrieval}

While the four scenarios described above differ in the form of the query, they share a common forensic characteristic: the need to retrieve candidate images under uncertainty and semantic abstraction. In practice, forensic experts often combine visual inspection with descriptive reasoning, iteratively refining searches based on both appearance and meaning. The unified retrieval framework proposed in this study mirrors this process by treating textual and visual representations as complementary sources of evidence rather than competing modalities.

By addressing photographic, textual, and sketch-based queries within a single methodological framework, this work aims to demonstrate the feasibility and value of multimodal retrieval in realistic forensic conditions. This unified perspective supports the broader goal of developing retrieval systems that are not only accurate but also aligned with the interpretative and evidentiary requirements of forensic science.

\section{Methodology overview}

\begin{figure}[htbp]
	\centering
	\includegraphics[page=1, width=1\textwidth]{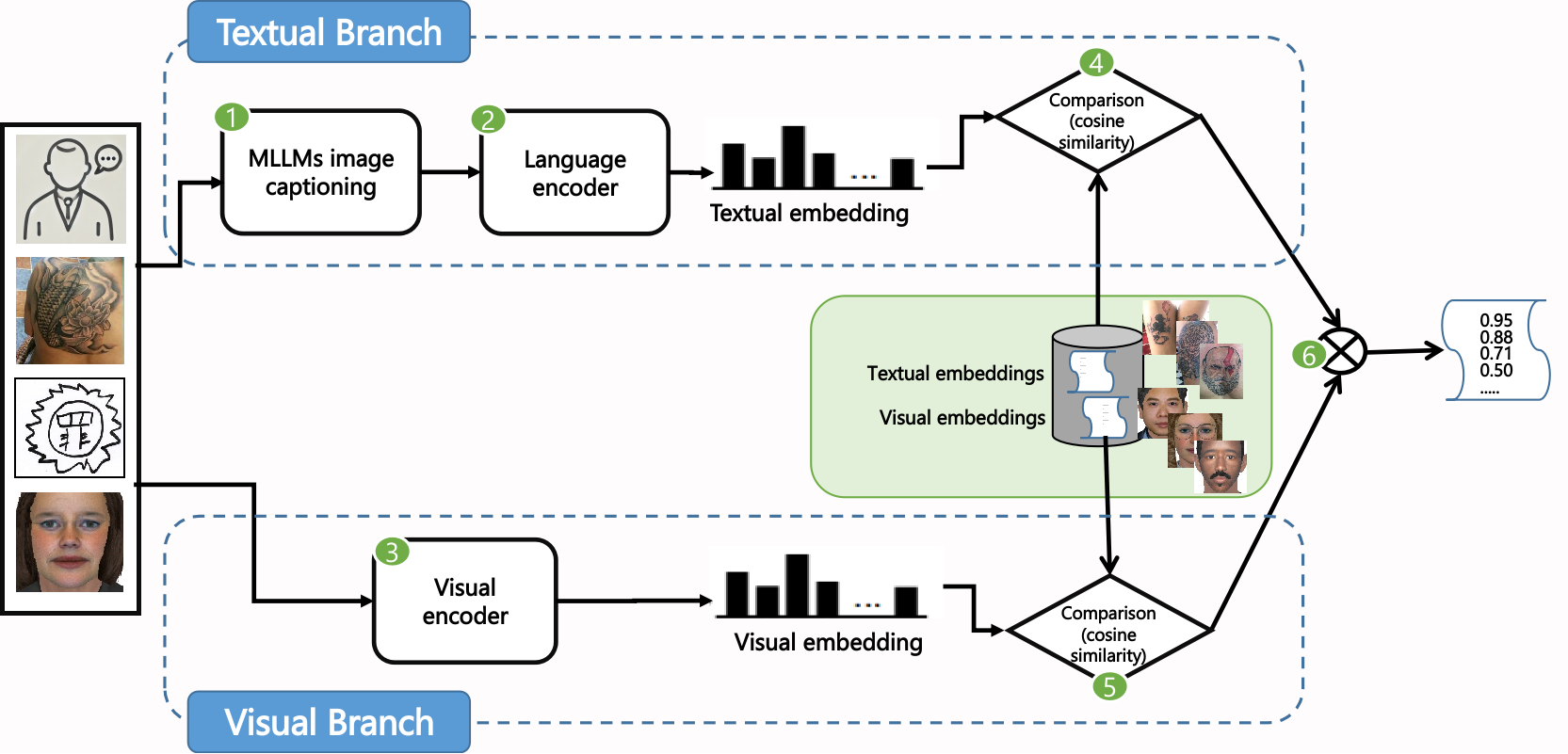}
	\caption{Overview of the proposed multimodal forensic retrieval framework, illustrating the visual and textual branches used to generate embeddings from different types of queries, including tattoo images, sketches, facial sketches, and textual descriptions. }
	\label{fig:proposal}
\end{figure}

The overall concept of the proposed approach is illustrated in Figure \ref{fig:proposal}. The framework is organized into two complementary branches: a visual branch and a textual branch. An input query—whether a tattoo described in text, a photographic tattoo image, a tattoo sketch, or a facial sketch—is processed through one or both branches to generate the corresponding embeddings. Visual queries are analyzed by the visual branch to produce visual feature embeddings, while all queries are processed by the textual branch to obtain semantic textual embeddings, with the exception of purely textual tattoo descriptions, which are handled exclusively by the textual branch. The resulting representations are then used for retrieval. The following sections describe each component of this process in greater detail.

\subsection{MLLM-based image captioning of query/gallery items  }   
\label{MLLM captions}
The first stage of the proposed framework consists of converting tattoo images, tattoo sketches and facial sketches into structured, interpretable, and machine-processable textual descriptions using Multimodal Large Language Models (MLLMs). This is depicted as Step 1 in Figure \ref{fig:proposal}. Unlike conventional visual-only approaches, this step explicitly captures the semantic content of tattoos through natural language, including symbolic motifs, stylistic elements, textual components, and broader thematic or narrative characteristics. Such semantic representations are particularly relevant in forensic contexts, where tattoos are often described and interpreted using language rather than precise visual measurements.

Given the extensive landscape of MLLMs reported in the literature \cite{CIRSurvey2025}, we selected five publicly available small- to medium-scale state-of-the-art models. This selection is grounded in 
previous work \cite{MLLMTattoo_CIARP2025}, where these specific architectures demonstrated superior retrieval performance. The selected models span different architectural designs and parameter scales, allowing us to assess the impact of model capacity and computational complexity on the quality of the generated descriptions:

\begin{itemize}
	\item \textbf{DeepSeek-VL2-tiny (1B)} \cite{DeepSeek-VL2}. A 1-billion-parameter vision–language model based on a Mixture-of-Experts (MoE) architecture with dynamic image tiling. It provides efficient multimodal understanding for tasks such as optical character recognition, visual question answering, and visual grounding, making it suitable for resource-constrained forensic environments.
	\item \textbf{Qwen2-VL-2B} \cite{Qwen2VL}. A lightweight 2-billion-parameter vision–language model supporting flexible image resolutions and robust cross-modal reasoning. It offers a favorable balance between descriptive accuracy and computational efficiency.
	\item \textbf{Bunny-v1.1-4B} \cite{Bunny2024}. A compact 4-billion-parameter multimodal model built on a SigLIP vision encoder and a Phi-3-mini language backbone, enhanced with an S-Wrapper architecture. Despite its moderate size, it achieves competitive performance on multimodal reasoning benchmarks due to high-quality training data.
	\item \textbf{Qwen2.5-VL-7B} \cite{Qwen2VL}. A 7-billion-parameter model with improved visual reasoning, precise spatial grounding, and advanced multimodal comprehension capabilities. Compared to its smaller counterpart, it offers higher descriptive accuracy at the cost of increased computational requirements.
	\item \textbf{LLaVA-1.6-7B} \cite{LLaVA}. A 7-billion-parameter multimodal model based on the Vicuna language backbone and a CLIP-based vision encoder, supporting higher input resolutions and enhanced visual reasoning through extended instruction tuning.
\end{itemize}

To ensure consistency and reproducibility, each model is prompted using standardized instructions designed to produce objective and semantically informative descriptions. We further investigate the influence of prompt specificity on the generated outputs by employing several levels of descriptive detail, adjusting in each case for tattoo or face images

We further investigate the influence of prompt specificity on the generated outputs by employing task-dependent prompts, tailored to the characteristics of tattoo images and face images, and designed to produce different levels of descriptive detail:

For tattoo images:
\begin{itemize}
	\item \textbf{Prompt 1}: “Can you describe this tattoo?” — a generic prompt aimed at producing a high-level description.
	\item \textbf{Prompt 2}: “Describe the elements present in this tattoo.” — a more focused prompt encouraging the identification of distinctive visual components.
	\item \textbf{Prompt 3}: “Describe the elements present in this tattoo. Focus on colors, shapes, styles, text, and meaningful objects.” — a domain-oriented prompt intended to capture detailed and forensically relevant attributes.
\end{itemize}

For face images:
\begin{itemize}
	\item \textbf{Prompt 1}: “Can you describe this face?” — a general prompt designed to produce an overall facial description.
	\item \textbf{Prompt 2}: “Describe the facial elements of this person.” — a prompt focused on anatomical and morphological characteristics.
	\item \textbf{Prompt 3}: “Describe the facial elements of this person. Focus on distinctive aspects like face feature dimensions, relations, race, gender, etc.” — a detailed prompt aimed at capturing attributes commonly used in forensic facial analysis.
\end{itemize}

This prompt design allows us to examine how varying semantic granularity and domain specificity influence the quality of the generated descriptions and, ultimately, retrieval performance across different forensic scenarios.

The proposed captioning strategy enhances operational transparency by providing reviewable attribute descriptions essential for legal admissibility. It also establishes a unified semantic space for heterogeneous forensic inputs, bridging visual evidence with human-centered reasoning where visual-only embeddings fall short

\subsection{Textual embeddings }

To generate textual embeddings (Step 2 in Figure~\ref{fig:proposal}), we employ the MPNet sentence transformer (\textit{all-mpnet-base-v2})~\cite{MPNet2020}. MPNet is specifically optimized for semantic representation learning and combines masked language modeling with permutation-based training objectives, allowing it to capture both contextual dependencies and global sentence-level meaning more effectively than earlier transformer architectures. In contrast to vision-language models such as CLIP, MPNet can encode substantially longer text sequences without aggressive truncation. This characteristic is particularly relevant in our setting, where MLLM-generated tattoo and face descriptions may vary considerably in length and level of detail. Its ability to preserve semantic information in extended descriptions makes it well suited for forensic retrieval tasks. Additionally, MPNet has demonstrated strong performance across multiple sentence embedding and semantic similarity benchmarks~\cite{ComparativeSentenceTranfs2025}. The model has also been successfully applied in 
previous work~\cite{MLLMTattoo_CIARP2025}, where it yielded competitive results in description-based retrieval scenarios. These empirical findings, together with its robustness to verbose inputs, support its selection as the textual backbone for our multimodal retrieval framework.

\subsection{Visual embeddings }
In addition to textual representations, the proposed framework incorporates task-specific visual feature extractors (Step 3 in Figure \ref{fig:proposal}) that have demonstrated strong performance in prior state-of-the-art studies. Given the domain-dependent nature of forensic image retrieval, we deliberately adopt visual descriptors that are well validated for tattoo recognition and facial sketch-based recognition, rather than introducing new visual models. This allows the study to focus on the impact of multimodal fusion while ensuring robust visual baselines for each forensic task.

 \subsubsection{Tattoo image visual features}

For tattoo image and sketch retrieval, we employ visual features commonly used in state-of-the-art tattoo recognition systems, as these representations have proven effective in capturing both low-level appearance and higher-level structural characteristics of tattoos. The following feature extractors are considered:

\begin{itemize}
	\item \textbf{MobileNetV2} \cite{Miguel_CIARP2019}. These features are extracted from the pool6 layer of a MobileNetV2 convolutional neural network pretrained on ImageNet. MobileNetV2 provides a compact and efficient representation while retaining sufficient discriminative power for tattoo recognition tasks.
	\item \textbf{Weighted Average Pooling (WAP)} \cite{Miguel_MTAP2022}. Built upon MobileNetV2 convolutional feature maps, WAP features introduce a weighted average pooling strategy that emphasizes salient tattoo regions. Local features are weighted using functions such as standard deviation, entropy, edge response, and skin masking, resulting in a more distinctive representation tailored to tattoo characteristics.
	\item \textbf{CLIP} \cite{CLIP_2021}. CLIP visual embeddings are 512-dimensional representations produced by CLIP’s image encoder (e.g., ViT or ResNet). These features capture high-level semantic content and are aligned with textual embeddings in a shared multimodal space, making them suitable for cross-modal retrieval and multimodal fusion.
	\item \textbf{TattTRN} \cite{Janier_TattTRN2024}. The Tattoo Template Reconstruction Network generates a dual-embedding representation by combining a raw image embedding, which captures low-level visual details, with a template-based embedding obtained through image-to-template translation. These complementary representations are concatenated into a unified 2,048-dimensional feature vector that emphasizes both appearance and core design structure.
\end{itemize}

 \subsubsection{Facial sketch visual features}

For facial sketch-based retrieval, we adopt visual feature extractors that have been specifically designed and validated for face recognition under challenging conditions, such as non-photographic inputs, low image quality, and high intra-class variability \cite{FaceFeaturesYoanna2021}. These characteristics closely match the properties of forensic facial sketches. The selected models are:

\begin{itemize}
	\item \textbf{MobileFaceNet} \cite{chen2018mobilefacenets}. An efficient convolutional neural network optimized for real-time face recognition on resource-constrained devices. MobileFaceNet employs residual bottleneck blocks with small expansion factors, replaces global average pooling with a Global Depth-wise Convolution (GDC) layer, and uses the PReLU activation function to enhance discriminative capacity.
	\item \textbf{ShuffleFaceNet} \cite{martindez2019shufflefacenet}. A lightweight face recognition architecture derived from ShuffleNetV2, designed to balance accuracy and computational efficiency. ShuffleFaceNet introduces architectural refinements and, similar to MobileFaceNet, employs PReLU activations and a GDC-based output layer to generate compact and discriminative facial embeddings.	
	\item \textbf{ResNet50 with ArcFace loss} \cite{deng2019arcface}. A deeper convolutional model that leverages the ResNet architecture for feature extraction. The use of the ArcFace loss function enhances inter-class separability and intra-class compactness, resulting in highly discriminative facial representations suitable for challenging forensic scenarios.
\end{itemize}

By employing these task-specific visual feature extractors, the proposed framework ensures that visual similarity is computed using representations that are well adapted to the unique characteristics of tattoos and facial sketches. This design choice allows for a meaningful evaluation of multimodal fusion strategies while maintaining strong and reliable visual baselines across all forensic retrieval scenarios.

\subsection{Score fusion: Combining textual and visual similarity}

Let $q_t$ and $q_v$ denote the textual and visual embeddings of the query, and let $g_t^i$ and $g_v^i$ denote the corresponding embeddings of the $i$-th gallery sample. Cosine similarity is computed independently in each modality (Steps 4 and 5 in Figure \ref{fig:proposal}):

\begin{equation}
	s_t^i = \cos(q_t, g_t^i), \qquad
	s_v^i = \cos(q_v, g_v^i).
\end{equation}

To obtain a unified identification score while preserving the zero-shot nature of the pipeline, we apply a parameter-free multiplicative fusion strategy (Step 6 in Figure \ref{fig:proposal}). This approach avoids the need for learning fusion weights on validation data, ensuring the system remains entirely training-free:

\begin{equation}
	S^i = s_t^i \cdot s_v^i.
\end{equation}

The final ranking is produced by sorting gallery samples in descending order of $S^i$.

This multiplicative fusion enforces cross-modal consistency by promoting candidates that achieve high similarity in both modalities simultaneously. In contrast, an additive weighted sum, may assign high scores to candidates that are strongly similar in only one modality while being weak in the other. The product formulation naturally penalizes such imbalance, as a low score in either modality proportionally reduces the final similarity. This behavior is particularly desirable in forensic retrieval, where agreement between semantic description and visual appearance increases evidential reliability.

\paragraph{Handling missing modalities.}
The framework supports unimodal retrieval when only one modality is available. If the query is text-only (e.g., witness description without an image), ranking is performed using $s_t^i$ alone. Similarly, if only a visual query is available, ranking relies exclusively on $s_v^i$. Therefore, multiplicative fusion is applied only in the multimodal setting, while unimodal scenarios reduce to standard cosine-based retrieval. This design ensures flexibility across operational scenarios while maintaining a simple and interpretable fusion mechanism.

\section{Experiments}
This section presents the empirical evaluation of the proposed multimodal retrieval framework. We assess its performance across the four distinct forensic scenarios designed to reflect real-world investigative challenges, ranging from standard image matching to heterogeneous queries involving natural language and sketches. The following subsections detail the experimental configuration, datasets, and metrics, followed by a comprehensive analysis of retrieval accuracy and operational implications.

\subsection{Experimental setup}
To ensure a proper assessment aligned with forensic standards, we establish a structured experimental environment. This setup encompasses the selection of relevant forensic datasets, the definition of performance metrics tailored to retrieval tasks, and the protocol for generating and processing textual descriptions. Each component is designed to validate the framework's robustness across different modalities and evidence types.

\subsubsection{Datasets: Forensic-relevant tattoo and face collections}

For experimental evaluation we used two public tattoo databases (BIVTatt~\cite{Miguel_MTAP2022} and WebTattoo~\cite{JointDetRecogWebTattoo2019}) and one public face sketch dataset (UoM-SGFS \cite{UoM-SGFSDataset}).  

\textbf{BIVTatt} contains 210 images representing 159 individuals, with some individuals contributing only a single image. The dataset includes 4,200 augmented images generated through 20 types of transformations. Following the protocol in~\cite{Miguel_MTAP2022}, we selected 16 transformations for our experiments, including variations in illumination, diffusion, affine transformations, rotations, and color changes. For identification, the probe set consisted of the transformed images, while the gallery set included only the original images. During testing, each probe’s source image was excluded from the gallery to avoid trivial matches.  

\textbf{WebTattoo} contains approximately 300,000 images across 600 tattoo categories collected from the web. For evaluation, the authors originally selected around 1,400 images from 400 classes for training and reserved images from 200 classes for testing. We used the publicly released training set, which contains 1,029 images from 399 individuals, without augmentation. This set was also used for the human-to-photo retrieval scenario, where we manually annotated textual descriptions for each tattoo. For the sketch-to-photo evaluation, we employed another set provided by WebTattoo as well, consisting of 300 sketch/photo pairs. Sketches were drawn by volunteers who observed a tattoo image for one minute and reproduced it the following day.  

\textbf{Face Sketch-to-Photo} experiments used the UoM-SGFS dataset, which contains 300 computer-generated facial composites created with the EFIT-V software, corresponding to 300 identities from the Color-FERET database. These composites were used in a 1-to-many identification protocol against the photo gallery.

\begin{figure}[htbp]
	\centering
	\includegraphics[width=1\linewidth]{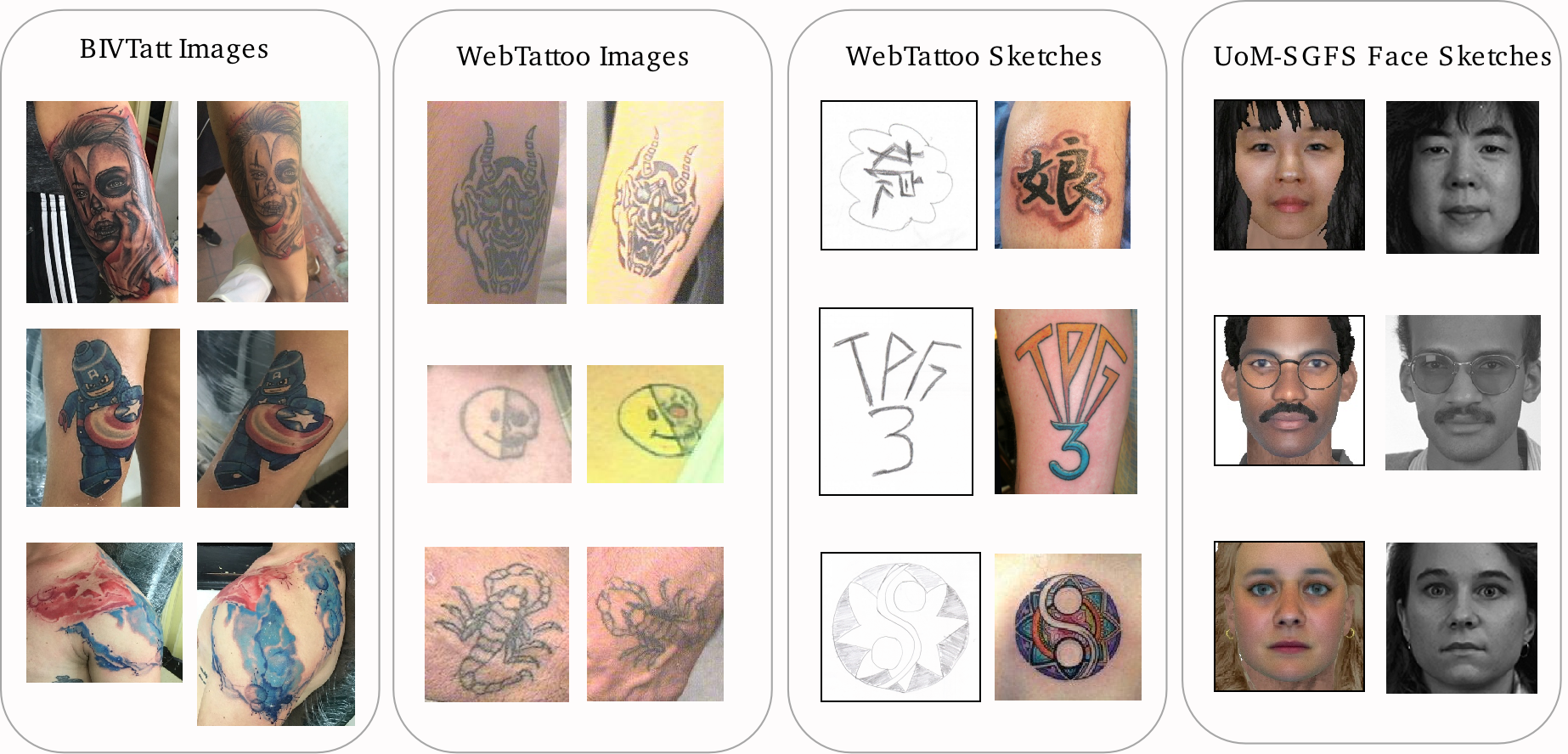}
	\caption{Samples of the four datasets employed in our experimental evaluation.}
	\label{fig:BD_samples.pdf}
\end{figure}

Figure \ref{fig:BD_samples.pdf} shows samples of the four previously described datasets in order to assess their characteristics and difficulty.

\subsubsection{Evaluation metrics and significance testing}

This section presents the quantitative evaluation of the proposed retrieval methods. We report standard ranking-based metrics that reflect both early retrieval performance and overall ranking quality. Statistical significance is assessed using bootstrap hypothesis testing, allowing us to determine whether observed differences between systems are likely due to chance.

Let $Q$ denote the number of queries. For each query $i \in \{1,\dots,Q\}$, two retrieval systems $A$ and $B$ return the rank of the first relevant item, denoted $r_i^A$ and $r_i^B$ (1-indexed).

\paragraph{Rank@$K$.}
Rank@$K$ measures the proportion of queries for which the relevant item appears within the top $K$ retrieved results. It reflects early retrieval performance, which is particularly important in forensic scenarios where investigators typically inspect only the top candidates.

For each query,
\begin{equation}
	s_i^A = \mathbbm{1}(r_i^A \le K), \qquad
	s_i^B = \mathbbm{1}(r_i^B \le K),
\end{equation}
and the aggregate metric is
\begin{equation}
	\text{Rank@}K(A) = \frac{1}{Q}\sum_{i=1}^{Q} s_i^A, \qquad
	\text{Rank@}K(B) = \frac{1}{Q}\sum_{i=1}^{Q} s_i^B,
\end{equation}
with observed difference
\begin{equation}
	\Delta_{\text{obs}} = \text{Rank@}K(A) - \text{Rank@}K(B).
\end{equation}

\paragraph{Mean Average Precision (mAP).}
Mean Average Precision evaluates the overall quality of the ranking by considering the exact position of the relevant item. Since each query has a single relevant item, the average precision for query $i$ reduces to
\begin{equation}
	\text{AP}_i = \frac{1}{r_i}.
\end{equation}
Thus,
\begin{equation}
	\text{mAP}(A) = \frac{1}{Q}\sum_{i=1}^{Q} \frac{1}{r_i^A},
\end{equation}
and similarly for $B$, with
\begin{equation}
	\Delta_{\text{obs}} = \text{mAP}(A) - \text{mAP}(B).
\end{equation}
Unlike Rank@$K$, mAP rewards systems that consistently rank the correct item as high as possible, even beyond a fixed cutoff.

\paragraph{Bootstrap hypothesis test.}
To assess whether performance differences are statistically significant, we tested the null hypothesis that both systems have equal expected performance. Let $d_i$ denote the per-query difference (either $s_i^A - s_i^B$ for Rank@$K$ or $\text{AP}_i^A - \text{AP}_i^B$ for mAP). The observed difference equals the sample mean of $\{d_i\}$.

Under the null hypothesis, the mean difference is zero. We therefore constructed centered differences
\begin{equation}
	\tilde{d}_i = d_i - \frac{1}{Q}\sum_{j=1}^{Q} d_j,
\end{equation}
and drew $T$ bootstrap samples with replacement from $\{\tilde{d}_i\}$. For each sample $t$, we computed the mean $\Delta_t^{(0)}$. The two-sided $p$-value was estimated as
\begin{equation}
	p = \frac{1}{T}\sum_{t=1}^{T}
	\mathbbm{1}\big(|\Delta_t^{(0)}| \ge |\Delta_{\text{obs}}|\big).
\end{equation}

\paragraph{Confidence intervals.}
To quantify uncertainty, we used the percentile bootstrap. Queries were resampled with replacement, the metric difference $\Delta_t$ was recomputed for each sample, and the $(1-\alpha)$ confidence interval was obtained from the empirical $\alpha/2$ and $1-\alpha/2$ quantiles.

\paragraph{Multiple comparisons.}
When multiple pairwise comparisons were conducted, we controlled the false discovery rate using the Benjamini–Hochberg procedure with $\alpha = 0.05$. Statistical significance was determined after this correction.

\subsubsection{Human textual description generation and processing}

To simulate realistic operational conditions, three forensic experts were asked to produce textual descriptions for each tattoo in the training subset of the WebTattoo dataset. The descriptions were written in Spanish, the experts’ native language, to ensure maximum descriptive precision and semantic richness.

Since the MPNet language encoder employed in our experiments is optimized for English, all descriptions were automatically translated into English using the Helsinki-NLP \textit{opus-mt-es-en} neural machine translation model.\footnote{Helsinki-NLP/opus-mt-es-en available on Hugging Face.} 

We also evaluated a multilingual variant of MPNet (\textit{paraphrase-multilingual-mpnet-base-v2})\footnote{sentence-transformers/paraphrase-multilingual-mpnet-base-v2 available on Hugging Face.} to avoid the translation step. However, this multilingual model consistently produced lower retrieval performance compared to the pipeline based on Spanish-to-English translation followed by the English MPNet encoder. Therefore, all reported results rely on the translated descriptions encoded with the English-based MPNet model.

\subsection{Results for tattoo photo-to-photo retrieval scenario}

We begin by evaluating the proposed framework on the Scenario 1 (photo-to-photo retrieval) using the WebTattoo and BIVTatt datasets. This scenario serves as a controlled baseline to assess the contribution of each component—prompts, visual features, and MLLM-generated descriptions—before addressing more heterogeneous forensic queries.

\paragraph{Baseline performance and multimodal fusion.}
We first establish a baseline using visual features alone (Table \ref{tab:photo_webtattoo_map} and \ref{tab:photo_bivtatt_map}, row \textit{base}). Interestingly, generic pre-trained models like \textit{CLIP} achieve strong standalone performance (mAP 0.932 on WebTattoo), often surpassing domain-specific features like TattTRN (0.918) when used in isolation. However, the true potential of the framework emerges when combining visual features with textual embeddings. 
Tables \ref{tab:photo_webtattoo_map} and \ref{tab:photo_bivtatt_map} highlight a critical distinction: while CLIP shows marginal or non-significant gains from added text (likely due to its existing image-text pre-training), domain-specific features (TattTRN and MobileNetV2) exhibit statistically significant improvements (marked with \ccheck) across most MLLM combinations. For instance, on WebTattoo, TattTRN improves from 0.918 to 0.963 mAP when paired with DeepSeek-VL2-tiny. This suggests that textual descriptions provide complementary semantic information that domain-specific visual models lack, whereas generic models already encode some of this alignment.

\begin{table}[htbp]
	\centering
	\caption{mAP results on WebTattooImages dataset. Significant improvements over base are marked with \ccheck{}.}
	\label{tab:photo_webtattoo_map}
	\begin{tabular}{lcccc}
		\toprule
		\textbf{Method} & \textbf{MobileNetV2} & \textbf{WAP} & \textbf{TattTRN} & \textbf{CLIP} \\
		\midrule
		base                 & 0.907\phantommark & 0.916\phantommark & 0.918\phantommark & 0.932\phantommark \\
		Bunny-v1.1-4B         & 0.951\ccheck      & 0.906\phantommark & 0.959\ccheck      & 0.931\phantommark \\
		DeepSeek-VL2-tiny & \textbf{0.963}\ccheck      & \textbf{0.923}\phantommark & \textbf{0.963}\ccheck      & \textbf{0.935}\phantommark \\
		Qwen2.5-VL-7B   & 0.931\phantommark & 0.845\phantommark & 0.951\ccheck      & 0.883\phantommark \\
		Qwen2-VL-2B          & 0.951\ccheck      & 0.883\phantommark & 0.954\ccheck      & 0.912\phantommark \\
		LLaVA-1.6-7B         & 0.934\phantommark & 0.853\phantommark & 0.941\ccheck      & 0.897\phantommark \\
		\bottomrule
	\end{tabular}%
\end{table}

\begin{table}[htbp]
	\centering
	\caption{mAP results on BIVTatt dataset. Significant improvements over base are marked with \ccheck{}.}
	\label{tab:photo_bivtatt_map}
	\begin{tabular}{lcccc}
		\toprule
		\textbf{Method} & \textbf{MobileNetV2} & \textbf{WAP} & \textbf{TattTRN} & \textbf{CLIP} \\
		\midrule
		base                 & 0.919\phantommark & 0.923\phantommark & 0.912\phantommark & 0.924\phantommark \\
		Bunny-v1.1-4B         & \textbf{0.964}\ccheck      & \textbf{0.953}\ccheck      & \textbf{0.960}\ccheck      & \textbf{0.959}\ccheck      \\
		DeepSeek-VL2-tiny & 0.957\ccheck      & 0.937\phantommark & 0.951\ccheck      & 0.950\ccheck      \\
		Qwen2.5-VL-7B   & 0.925\phantommark & 0.889\phantommark & 0.931\ccheck      & 0.905\phantommark \\
		Qwen2-VL-2B          & 0.935\phantommark & 0.908\phantommark & 0.940\ccheck      & 0.925\phantommark \\
		LLaVA-1.6-7B         & 0.931\phantommark & 0.893\phantommark & 0.926\ccheck      & 0.916\phantommark \\
		\bottomrule
	\end{tabular}%
\end{table}

\paragraph{Prompt selection.}
To determine the optimal textual description strategy, we evaluated three prompt variations while keeping visual and encoding parameters fixed. As shown in Table \ref{tab:prompt_comparison}, Prompt 2 yields statistically superior mAP scores, significantly outperforming Prompt 1 and Prompt 3 in 12 out of 20 pairwise comparisons. This suggests that the specific phrasing and structural constraints of Prompt 2 better elicit the semantic details relevant for tattoo discrimination. Consequently, Prompt 2 is used for all remaining experiments.

\begin{table}[htbp]
	\centering
	\caption{Pairwise comparison wins between prompts. Cell $(i,j)$ indicates how many times prompt $i$ significantly outperformed prompt $j$ from the $10$ comparisons.}
	\label{tab:prompt_comparison}
	\begin{tabular}{cccc}
		\toprule
		& \textbf{Prompt 1} & \textbf{Prompt 2} & \textbf{Prompt 3} \\
		\midrule
		\textbf{Prompt 1} & - & 0 & 9 \\
		\textbf{Prompt 2} & 4 & - & 8 \\
		\textbf{Prompt 3} & 3 & 0 & - \\
		\bottomrule
	\end{tabular}
\end{table}

\paragraph{Comparison of visual backbones in multimodal settings.}
To determine which visual feature benefits most from the multimodal framework, we performed a pairwise comparison of visual backbones across various prompt and MLLM configurations. Table \ref{tab:photo_pairwise_visual} aggregates these results, where each cell \textit{(i,j)} indicates how many times configurations based on feature \textit{i} significantly outperformed those based on feature j across 10 experimental variations.

\begin{table}[htbp]
	\centering
	\caption{Pairwise comparison wins between base methods. Cell $(i,j)$ indicates how many times method $i$ significantly outperformed method $j$ from the $10$ comparisons.}
	\label{tab:photo_pairwise_visual}
	\begin{tabular}{lcccc}
		\toprule
		\textbf{Method} & \textbf{MobileNetV2} & \textbf{WAP} & \textbf{TattTRN} & \textbf{CLIP} \\
		\midrule
		MobileNetV2      & --- & 10 & 1  & 10 \\
		WAP        & 0   & --- & 0  & 0  \\
		TattTRN    & 5   & 10 & --- & 10 \\
		CLIP& 0   & 10 & 0  & --- \\
		\bottomrule
	\end{tabular}
	
\end{table}

The results confirm that TattTRN and MobileNetV2 are the most effective backbones for this forensic task when augmented with text. TattTRN configurations dominate CLIP and WAP in all 10 comparisons, and win 5 out of 10 against MobileNetV2. This reinforces that domain-adapted visual features, when enriched with semantic descriptions, outperform generic visual-language models in specialized forensic retrieval.

\paragraph{MLLM performance and optimal configurations.}
Regarding the language models, a pairwise analysis (Table \ref{tab:photo_pairwise_mllm}) reveals that Bunny-v1.1-4B and DeepSeek-VL2-Tiny produce the most discriminative textual embeddings, significantly outperforming larger models like Qwen2.5-VL-7B or LLaVA-1.6-7B. This suggests that for forensic retrieval, the quality and relevance of the generated description are more critical than the parameter count of the language model.
Table \ref{tab:mllm_combinations_results_3dec} summarizes the top-performing configurations. The highest mAP scores (0.963) are achieved by combining DeepSeek-VL2-Tiny (with Prompt 2) and TattTRN. 

\begin{table}[htbp]
	\centering
	\caption{Pairwise significant wins between MLLM methods. Cell $(i,j)$ shows how many times method $i$ significantly outperformed method $j$. Diagonal is omitted.}
	\label{tab:photo_pairwise_mllm}
		\begin{tabular}{lccccc}
			\toprule
			\textbf{Method} & \textbf{Bunny-v1.1-4B} & \textbf{DeepSeek-VL2-tiny} & \textbf{LLaVA-1.6-7B} & \textbf{Qwen2-VL-2B} & \textbf{Qwen2.5-VL-7B} \\
			\midrule
			Bunny-v1.1-4B             & ---   & 4  & 8  & 7  & 7  \\
			DeepSeek-VL2-tiny & 1     & --- & 8  & 7  & 8  \\
			LLaVA-1.6-7B             & 0     & 0  & --- & 0  & 2  \\
			Qwen2-VL-2B              & 0     & 0  & 5  & --- & 7  \\
			Qwen2.5-VL-7B   & 0     & 0  & 1  & 0  & --- \\
			\bottomrule
		\end{tabular}%
\end{table}

\begin{table}[htpb]
	\centering
	\caption{Best combinations of methods for dataset WebTattoo. For clarity, only the highest-performing combinations are included, sorted in descending order of mAP.}
	\label{tab:mllm_combinations_results_3dec}
		\begin{tabular}{lccccc}
			\toprule
			\textbf{Method} & \textbf{mAP} & \textbf{Rank@1} & \textbf{Rank@10} & \textbf{Rank@20} & \textbf{Rank@50} \\
			\midrule
			DeepSeek-VL2-tiny\_Prompt2+TattTRN & \textbf{0.963} & \textbf{0.868} & 0.960 & 0.975 & 0.989 \\
			DeepSeek-VL2-tiny\_Prompt1+TattTRN & \textbf{0.963} & \textbf{0.868} & 0.954 & 0.980 & \textbf{0.993} \\
			DeepSeek-VL2-tiny\_Prompt3+TattTRN& \textbf{0.963} & 0.865 & 0.958 & \textbf{0.982} & 0.992 \\
			DeepSeek-VL2-tiny\_Prompt2+MobileNetV2   & \textbf{0.963} & 0.854 & \textbf{0.964} & 0.981 & 0.988 \\
			Bunny-v1.1-4B\_Prompt1+TattTRN             & 0.962 & 0.865 & 0.960 & 0.978 & 0.989 \\
			DeepSeek-VL2-tiny\_Prompt3+MobileNetV2   & 0.961 & 0.848 & 0.958 & 0.973 & 0.989 \\
			Bunny-v1.1-4B\_Prompt1+MobileNetV2               & 0.960 & 0.854 & 0.961 & 0.972 & 0.987 \\
			Bunny-v1.1-4B\_Prompt3+TattTRN             & 0.960 & 0.853 & 0.954 & 0.975 & 0.992 \\
			DeepSeek-VL2-tiny\_Prompt1+MobileNetV2   & 0.959 & 0.857 & 0.955 & 0.975 & 0.986 \\
			Bunny-v1.1-4B\_Prompt2+TattTRN             & 0.959 & 0.857 & 0.953 & 0.976 & 0.988 \\
			\bottomrule
		\end{tabular}%
\end{table}

\begin{figure}[htb]
	\centering
	\includegraphics[width=0.8\linewidth]{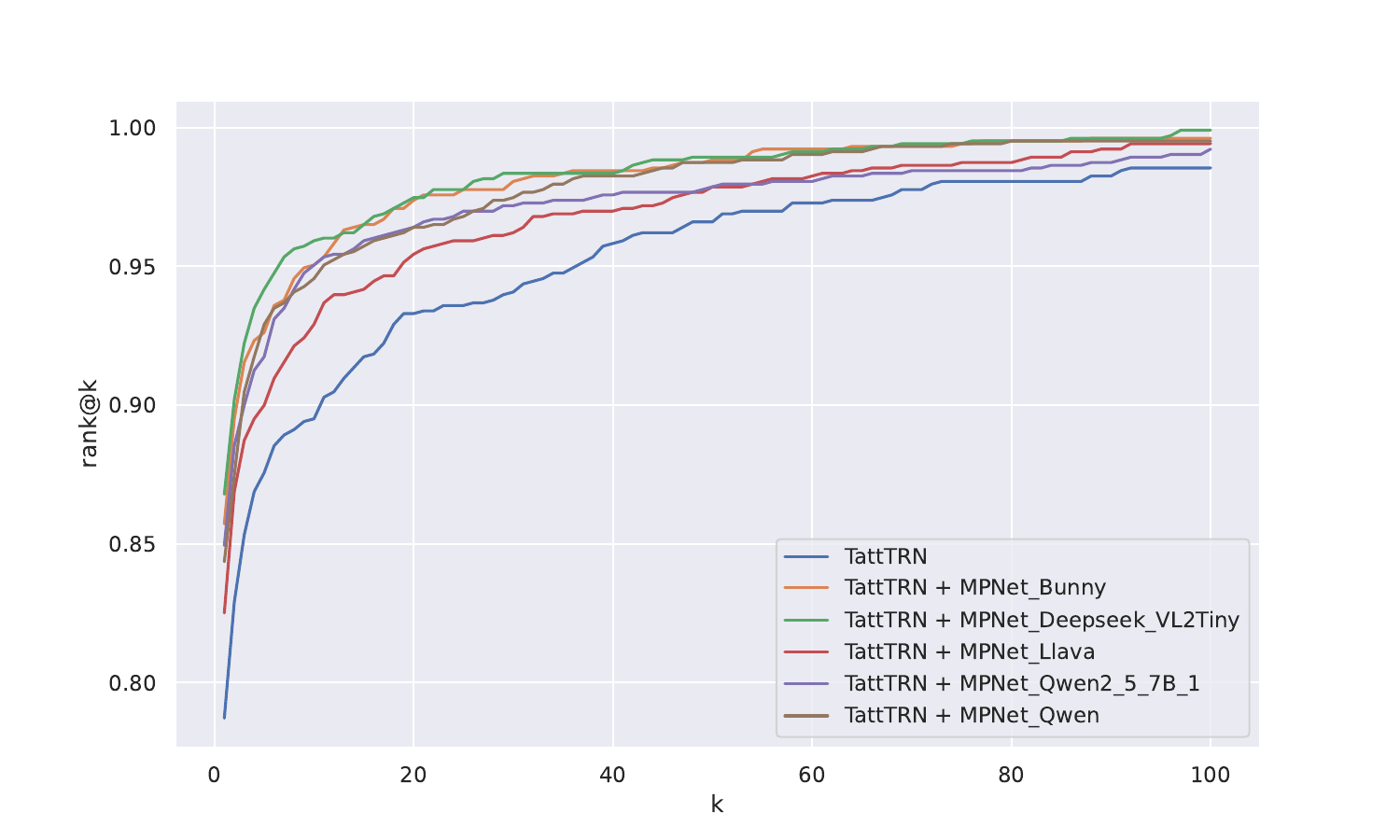}
	\caption{Rank@K performance comparison on WebTattoo. Combining TattTRN visual features with MLLM-generated textual descriptions consistently improves retrieval accuracy across all ranks.}
	\label{fig:photo_rank_at_k_webtattoo}
\end{figure}

From an operational forensic perspective, the improvement in \textbf{Rank@1} is particularly critical: the best multimodal configuration raises the probability of retrieving the correct match as the top result from 0.82 (visual-only) to 0.87. Figure \ref{fig:photo_rank_at_k_webtattoo} further illustrates that this performance gain is consistent across all rank thresholds.

In summary, these results validate the core hypothesis: integrating semantically rich textual descriptions with domain-adapted visual features creates a more robust representation for forensic retrieval. The synergy between TattTRN and compact MLLMs like DeepSeek-VL2-Tiny provides a significant advantage over unimodal approaches, laying a strong foundation for the heterogeneous scenarios evaluated in the following sections.

\subsection{Results for tattoo human description-to-photo retrieval scenario}
\label{sec:exp_human_description}
This subsection evaluates Scenario 2, which is a critical operational configuration: retrieval driven exclusively by textual descriptions provided by human users (e.g., witness statements or expert reports). In this setting, no visual query data is available. The gallery images are encoded using descriptions generated by different MLLMs, while user queries are encoded using the same language encoder to ensure modality compatibility. This experiment isolates the effectiveness of the semantic space in bridging human language and visual evidence.

\paragraph{Prompt selection.}
We first analyze the effect of prompt formulation on retrieval performance. Pairwise statistical comparisons were conducted using the mAP metric across the three evaluated prompts. The results, summarized in Table \ref{tab:human_compare_by_promt}, indicate that differences among prompts are generally limited in this text-only setting. Nevertheless, Prompt 1 exhibits a slight but consistent advantage, achieving a higher number of statistical wins (5 out of 15 comparisons) compared to Prompt 2 and Prompt 3. Based on this, Prompt 1 is selected as the default configuration for subsequent analyses, although Prompt 2 remains highly competitive.

\begin{table}[htbp]
	\centering
	\caption{Pairwise comparison between prompts for human description queries. Each cell contains the number of times the prompt in the row was significantly superior to the prompt in the column (out of 15 comparisons).}
	\label{tab:human_compare_by_promt}
	\begin{tabular}{cccc}
		\toprule
		& \textbf{Prompt 1} & \textbf{Prompt 2} & \textbf{Prompt 3} \\
		\midrule
		\textbf{Prompt 1} & --- & 2 & 3 \\
		\textbf{Prompt 2} & 2 & --- & 0 \\
		\textbf{Prompt 3} & 1 & 1 & --- \\
		\bottomrule
	\end{tabular}
\end{table}

\paragraph{Impact of human annotation variability.}
A crucial factor in forensic workflows is the consistency of human input. We examined the influence of annotation quality by comparing performance across descriptions provided by three different human expert annotators. Pairwise comparisons over mAP (Table \ref{tab:human_compare_by_human}) reveal substantial performance differences. \textbf{Expert 1} consistently underperforms, losing all 15 comparisons against Expert 2 and 13 against Expert 3. Conversely, Experts 2 and 3 show comparable performance, with Expert 2 holding a slight edge. 
This finding underscores that \textit{who} provides the description is as critical as the retrieval system itself. Variability in annotation style, vocabulary, and detail level can outweigh the effects of prompt choice or model selection. For operational deployment, this suggests a need for standardized description protocols or training for personnel generating forensic queries.

\begin{table}[htbp]
	\centering
	\caption{Pairwise comparison between human expert annotators. Each cell indicates the number of times the expert in the row significantly outperformed the expert in the column (out of 15 comparisons).}
	\label{tab:human_compare_by_human}
	\begin{tabular}{cccc}
		\toprule
		& \textbf{Expert 1} & \textbf{Expert 2} & \textbf{Expert 3} \\
		\midrule
		\textbf{Expert 1} & --- & 0 & 0 \\
		\textbf{Expert 2} & 15 & --- & 3 \\
		\textbf{Expert 3} & 13 & 0 & --- \\
		\bottomrule
	\end{tabular}
\end{table}

\paragraph{MLLM encoder performance.}
We then compared the different MLLMs employed to encode the gallery descriptions. The pairwise results (Table \ref{tab:human_compare_by_mllm}) show that DeepSeek-VL2-Tiny and Bunny-v1.1-4B consistently outperform the remaining models, achieving the highest number of wins. DeepSeek-VL2-tiny, in particular, dominates the comparison, winning 9 out of 9 tests against LLaVA-1.6-7B and Qwen2.5-VL-7B. In contrast, larger models like Qwen2.5-VL-7B and LLaVA-1.6-7B generally obtain lower performance in this text-only retrieval setting. This reinforces the observation from the photo-to-photo scenario: smaller models often generate more discriminative embeddings for forensic tasks than larger, general-purpose models.
\begin{table}[htbp]
	\centering
	\caption{Pairwise comparison between MLLM encoding methods. Each cell shows the number of times the MLLM in the row significantly outperformed the MLLM in the column (out of 9 comparisons).}
	\label{tab:human_compare_by_mllm}
	\begin{tabular}{lccccc}
		\toprule
		& \textbf{Bunny-v1.1-4B} & \textbf{DeepSeek-VL2-tiny} & \textbf{LLaVA-1.6-7B} & \textbf{Qwen2-VL-2B} & \textbf{Qwen2.5-VL-7B} \\
		\midrule
		\textbf{Bunny-v1.1-4B} & --- & 0 & 9 & 4 & 8 \\
		\textbf{DeepSeek-VL2-tiny} & 0 & --- & 9 & 6 & 9 \\
		\textbf{LLaVA-1.6-7B} & 0 & 0 & --- & 0 & 0 \\
		\textbf{Qwen2-VL-2B} & 0 & 0 & 8 & --- & 0 \\
		\textbf{Qwen2.5-VL-7B} & 0 & 0 & 2 & 6 & --- \\
		\bottomrule
	\end{tabular}
\end{table}

\paragraph{Optimal configurations and operational relevance.}
Table \ref{tab:tattoo_human_comparison} presents the top-performing parameter combinations ranked by mAP. The best configuration (DeepSeek-VL2-tiny\_Prompt2\_Expert2) achieves a mAP of 0.745. Notably, while Prompt 1 was selected for general robustness, the absolute peak performance was achieved with Prompt 2 when paired with Expert 2, highlighting the sensitivity of the system to specific prompt-expert interactions.

\begin{table}[htbp]
	\centering
	\caption{Performance comparison of the top parameter configurations for human description-to-photo retrieval according to mAP and Rank@K metrics. For clarity, only the highest-performing combinations are included, sorted in descending order of mAP.}
	\label{tab:tattoo_human_comparison}
	\begin{tabular}{lrrrrr}
		\toprule
		Method & map & rank1 & rank10 & rank20 & rank50 \\
		\midrule
		DeepSeek-VL2-tiny\_Prompt2\_Expert2 & \textbf{0.745} & \textbf{0.429} & 0.689 & 0.759 & \textbf{0.862} \\
		DeepSeek-VL2-tiny\_Prompt1\_Expert2 & 0.738 & \textbf{0.429} & 0.682 & \textbf{0.767} & 0.857 \\
		DeepSeek-VL2-tiny\_Prompt3\_Expert2 & 0.734 & 0.406 & 0.694 & 0.752 & 0.850 \\
		Bunny-v1.1-4B\_Prompt2\_Expert2    & 0.723 & 0.419 & \textbf{0.697} & 0.747 & 0.825 \\
		DeepSeek-VL2-tiny\_Prompt1\_Expert3 & 0.723 & 0.393 & 0.644 & 0.737 & 0.850 \\
		DeepSeek-VL2-tiny\_Prompt3\_Expert3 & 0.722 & 0.356 & 0.664 & 0.762 & 0.852 \\
		Qwen2-VL-2B\_Prompt2\_Expert2     & 0.722 & 0.373 & 0.667 & 0.744 & 0.845 \\
		Bunny-v1.1-4B\_Prompt3\_Expert2    & 0.721 & 0.388 & 0.639 & 0.737 & 0.852 \\
		DeepSeek-VL2-tiny\_Prompt2\_Expert3 & 0.711 & 0.373 & 0.647 & 0.734 & 0.840 \\
		\bottomrule
	\end{tabular}
\end{table}

Overall, these results show that reliable tattoo retrieval is feasible even in the absence of visual information, relying exclusively on textual descriptions. The best configurations achieve competitive mAP and strong early-rank accuracy, indicating that semantically rich LLM-generated descriptions can effectively bridge the gap between human queries and image representations. However, the Rank@1 scores ($\approx 0.43$) are notably lower than in the photo-to-photo scenario (0.87). This gap reflects the inherent ambiguity and subjective variability in human language compared to visual data. While the system successfully narrows down the search space ($Rank@10 \approx 0.69$), operational workflows should anticipate the need for manual review of top candidates rather than relying on a single automatic match. This validates the approach as a powerful investigative lead generator, particularly in cases where visual evidence is unavailable.

\subsection{Results for tattoo sketch-to-photo retrieval scenario}
\label{sec:exp_sketch_photo}

This subsection addresses Scenario 3: the heterogeneous sketch-to-photo retrieval scenario, which represents one of the most challenging forensic tasks due to the significant modality gap between hand-drawn sketches and photographic evidence. In this setting, query sketches are encoded using both visual and textual features, as well as the gallery, which consists of photographic tattoo images.

\paragraph{Baseline performance and visual features.}

We first establish the baseline performance using visual features alone (Table \ref{tab:sketch_visualmllm_map}, row \textit{base}). In contrast to the photo-to-photo scenario, generic models like CLIP struggle significantly in this cross-modal setting (mAP 0.597), performing only marginally better than standard CNN backbones (MobileNetV2 $\approx$ 0.54). However, the domain-specific TattTRN feature demonstrates remarkable robustness, achieving a standalone mAP of 0.816. 
A pairwise statistical comparison (Table \ref{tab:sketch_visual_pairwise_wins}) confirms the superiority of TattTRN in this context. Configurations based on TattTRN significantly outperform all other visual backbones in every comparison (4 out of 4 wins against each competitor). This performance gap can be attributed to TattTRN's unique training methodology. As described by \cite{Janier_TattTRN2024}, TattTRN was trained on a semi-synthetic dataset where predefined tattoo templates (structurally similar to sketches) were blended onto real skin images from the NTU databases with augmentations such as Gaussian blur and opacity reduction. This exposure to sketch-like structures overlaid on realistic textures during training effectively pre-adapts the model to the modality gap encountered in forensic sketch retrieval, allowing it to capture structural patterns invariant to drawing style that generic features miss.

\begin{table}[htbp]
	\centering
	\caption{Pairwise significant wins between base visual methods on WebTattoo\_Sketch. Cell $(i,j)$ shows how many times method $i$ significantly outperformed method $j$.}
	\label{tab:sketch_visual_pairwise_wins}
		\begin{tabular}{lcccc}
			\toprule
			\textbf{Method} & \textbf{MobileNetV2} & \textbf{WAP} & \textbf{TattTRN} & \textbf{CLIP} \\
			\midrule
			MobileNetV2 & ---   & 0 & 0 & 0 \\
			WAP   & 2     & --- & 0 & 0 \\
			TattTRN           &  4     &  4  & --- &  4 \\
			CLIP              & 1     & 0  & 0  & --- \\
			\bottomrule
		\end{tabular}%
\end{table}

\paragraph{Impact of multimodal fusion.}

We then evaluated the benefit of augmenting visual features with MLLM-generated textual descriptions (Table \ref{tab:sketch_visualmllm_map}). The results show statistically significant improvements (marked with \ccheck) across almost all combinations. For instance, adding textual embeddings to TattTRN increases mAP from 0.816 to 0.888 (DeepSeek-VL2-tiny). Even generic features like CLIP see substantial gains (0.597 to 0.789), suggesting that semantic descriptions provide a stable common ground when visual features are compromised by the modality gap. Nevertheless, the absolute performance ceiling is still defined by the visual backbone, with TattTRN combinations consistently leading the rankings.

\begin{table}[htbp]
	\centering
	\caption{mAP results on WebTattoo\_Sketch dataset. Significant improvements over base are marked with \ccheck{}.}
	\label{tab:sketch_visualmllm_map}
	\begin{tabular}{lcccc}
		\toprule
		\textbf{Method} & \textbf{MobileNetV2} & \textbf{WAP} & \textbf{TattTRN} & \textbf{CLIP} \\
		\midrule
		base                 & 0.538\phantommark & 0.549\phantommark & 0.816\phantommark & 0.597\phantommark \\
		Bunny-v1.1-4B         & 0.738\ccheck      & 0.766\ccheck      & 0.872\ccheck      & 0.782\ccheck      \\
		DeepSeek-VL2-tiny & \textbf{0.760}\ccheck      & \textbf{0.795}\ccheck      & \textbf{0.888}\ccheck      & 0.789\ccheck      \\
		Qwen2-VL-2B          & 0.753\ccheck      & 0.777\ccheck      & 0.884\ccheck      & \textbf{0.791}\ccheck      \\
		Qwen2.5-VL-7B   & 0.711\ccheck      & 0.721\ccheck      & 0.874\phantommark & 0.737\ccheck      \\
		\bottomrule
	\end{tabular}%
\end{table}

\paragraph{Prompt and MLLM analysis.}
Regarding prompt formulation, the results indicate higher stability compared to the human description scenario. Pairwise comparisons show that Prompt 2 is only significantly superior in one instance against Prompts 1 and 3. This suggests that when describing sketches (which may contain less subjective variability than witness testimony), the specific prompt phrasing has a diminished impact on retrieval performance.
In terms of MLLM selection, Table \ref{tab:sketch_pairwise_textual} presents the pairwise wins for textual encoders paired with TattTRN. DeepSeek-VL2-tiny emerges as the most robust option, significantly outperforming Qwen2.5-VL-7B in 3 out of 4 comparisons and Bunny-v1.1-4B in 1 out of 4. While Bunny-v1.1-4B and Qwen2-VL-2B remain competitive, DeepSeek-VL2-tiny consistently secures the top positions in combined metrics.

\begin{table}[htpb]
	\centering
	\caption{Pairwise significant wins between MLLM methods on WebTattoo\_Sketch. Cell $(i,j)$ shows how many times method $i$ significantly outperformed method $j$.}
	\label{tab:sketch_pairwise_textual}
		\begin{tabular}{lcccc}
			\toprule
			\textbf{Method} & \textbf{Bunny-v1.1-4B} & \textbf{DeepSeek-VL2-tiny} & \textbf{Qwen2-VL-2B} & \textbf{Qwen2.5-VL-7B} \\
			\midrule
			Bunny-v1.1-4B             & ---   & 0  & 0  & 0  \\
			DeepSeek-VL2-tiny & 1     & --- & 0  & 3  \\
			Qwen2-VL-2B              & 0     & 0  & --- & 2  \\
			Qwen2.5-VL-7B   & 0     & 0  & 0  & --- \\
			\bottomrule
		\end{tabular}%
\end{table}

\paragraph{Optimal configurations and operational relevance.}
Table \ref{tab:sketch_best_results} summarizes the top-performing configurations. All top-10 entries utilize the TattTRN visual feature, reinforcing its necessity for sketch-based tasks. The highest mAP (0.898) is achieved by DeepSeek-VL2-Tiny with Prompt 3, while the best Rank@1 (0.649) is obtained with Prompt 2. 

\begin{table}[htbp]
	\centering
	\caption{Best evaluation results for MLLM+TattTRN combinations. Bold values indicate the best per column.}
	\label{tab:sketch_best_results}
		\begin{tabular}{lccccc}
			\toprule
			\textbf{Method} & \textbf{mAP} & \textbf{Rank@1} & \textbf{Rank@10} & \textbf{Rank@20} & \textbf{Rank@50} \\
			\midrule
			DeepSeek-VL2-tiny\_Prompt3+TattTRN & \textbf{0.898} & 0.622 & \textbf{0.892} & 0.926 & \textbf{0.973} \\
			DeepSeek-VL2-tiny\_Prompt2+TattTRN & 0.888 & \textbf{0.649} & 0.858 & 0.932 & \textbf{0.973} \\
			Qwen2-VL-2B\_Prompt3+TattTRN              & 0.888 & 0.588 & 0.872 & \textbf{0.939} & 0.966 \\
			Qwen2-VL-2B\_Prompt3+TattTRN              & 0.885 & 0.622 & 0.865 & 0.926 & \textbf{0.973} \\
			Qwen2-VL-2B\_Prompt2+TattTRN              & 0.884 & 0.581 & 0.872 & 0.919 & 0.966 \\
			Bunny-v1.1-4B\_Prompt3+TattTRN             & 0.883 & 0.547 & 0.878 & 0.932 & \textbf{0.973} \\
			DeepSeek-VL2-tiny\_p1+TattTRN & 0.881 & 0.595 & 0.878 & 0.912 & 0.959 \\
			Qwen2.5-VL-7B\_Prompt2+TattTRN   & 0.874 & 0.568 & 0.851 & 0.912 & 0.966 \\
			Bunny-v1.1-4B\_Prompt2+TattTRN             & 0.872 & 0.568 & 0.851 & 0.899 & 0.966 \\
			Qwen2.5-VL-7B\_Prompt1+TattTRN   & 0.871 & 0.581 & 0.851 & 0.912 & \textbf{0.973} \\
			\bottomrule
		\end{tabular}%
\end{table}

\begin{figure}[htbp]
	\centering
	\includegraphics[width=0.8\linewidth]{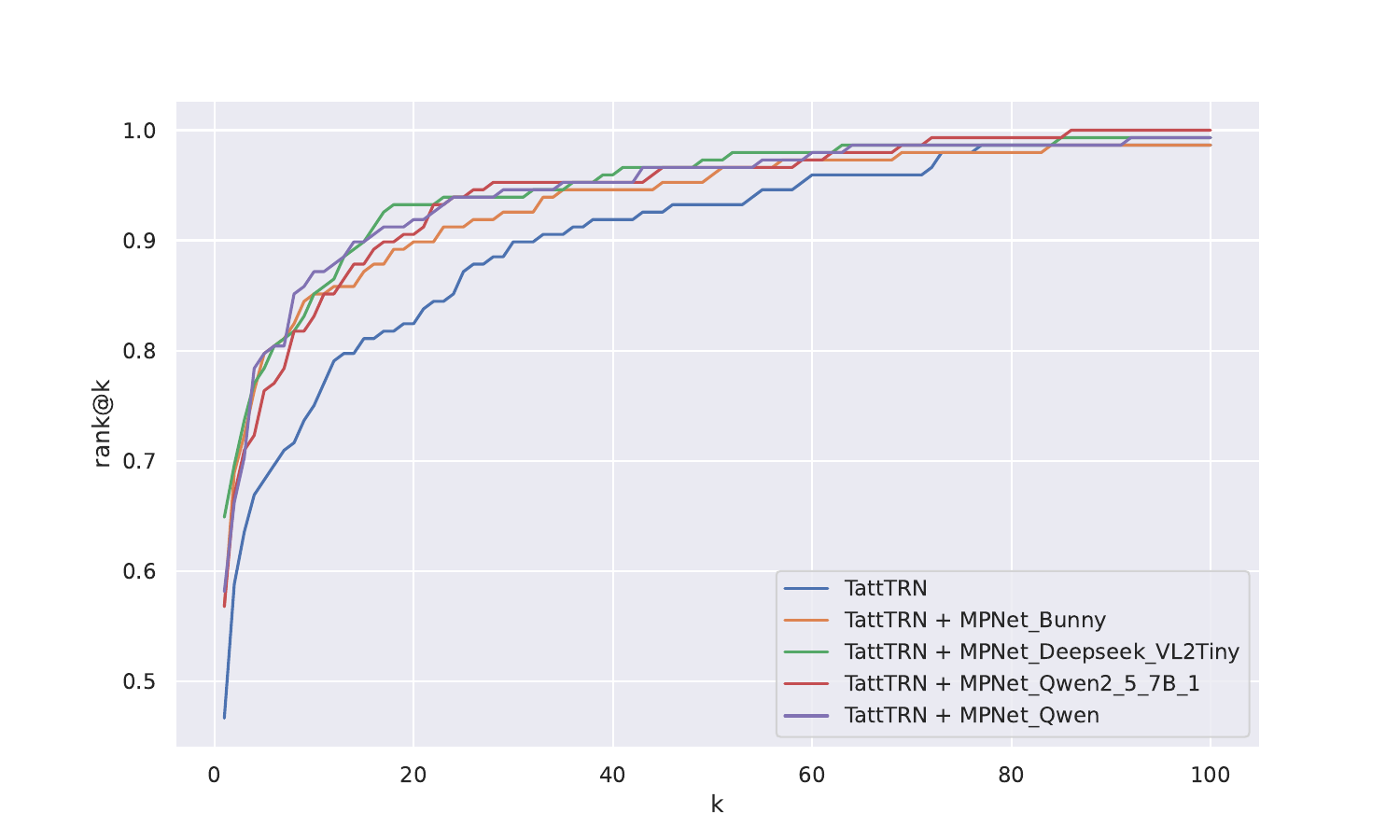}
	\caption{Rank@K performance comparison on the Sketch dataset. Combining TattTRN visual features with MLLM-generated textual descriptions improves retrieval accuracy across all ranks.}
	\label{fig:rank_at_k_sketch}
\end{figure}

Figure \ref{fig:rank_at_k_sketch} illustrates the performance gain across rank thresholds. The multimodal approach consistently outperforms the visual-only baseline at every level, ensuring that relevant matches are not pushed deeper into the retrieval list. 

In summary, the sketch-to-photo scenario confirms that while domain-specific visual features (TattTRN) are indispensable for handling the modality gap, semantic textual descriptions provide a crucial complementary signal. The best configuration achieves a Rank@1 of 0.649, meaning that in nearly 65 \% of cases, the correct match is identified immediately despite the inherent ambiguity of sketch evidence. It is important to note, however, that the evaluation sketches in the WebTattoo dataset were produced by experts viewing the source image, representing a controlled modality gap. In operational scenarios, sketches derived from witness memory may exhibit greater visual distortion, potentially diminishing the effectiveness of visual-only features. Under such conditions, the semantic component would likely assume an even more critical role in maintaining retrieval accuracy. Consequently, this performance level supports the operational viability of the framework for assisting forensic artists and investigators in narrowing down suspect pools from sketch evidence.

\subsection{Results for face sketch-to-photo retrieval scenario}
\label{sec:exp_face_sketch}
This subsection evaluates Scenario 4 (face sketch-to-photo retrieval scenario), a classical heterogeneous recognition problem in forensic biometrics. Unlike tattoo retrieval, where semantic content varies widely, facial sketches often adhere to more standardized anatomical structures, potentially altering the utility of textual descriptions. In this setting, query sketches are encoded using MLLM-generated descriptions, while the gallery consists of photographic face images encoded with visual and textual features.

\paragraph{Baseline performance and visual features.}
We first establish the baseline performance using visual features alone. Table \ref{tab:face_compare_by_visual_encoder} summarizes pairwise statistical comparisons across 10 experimental variations. ResNet50 emerges as the dominant visual backbone, significantly outperforming both MobileFaceNet and ShuffleFaceNet in all comparisons (18 out of 20 wins). This establishes a strong visual baseline (mAP 0.353), indicating that for facial data, standard deep learning backbones trained on large-scale face recognition tasks retain significant discriminative power even when matching against sketches. This contrasts with the tattoo sketch scenario, where generic models struggled significantly against domain-specific features.

\begin{table}[htbp]
	\centering
	\caption{Pairwise statistical comparison between visual encoders for face sketch retrieval. Each cell reports the number of times (out of 10 comparisons) that the encoder in the row was significantly superior to the encoder in the column.}
	\label{tab:face_compare_by_visual_encoder}
	\begin{tabular}{cccc}
		\toprule
		& \textbf{ResNet50} & \textbf{MobileFaceNet} & \textbf{ShuffleFaceNet} \\
		\midrule
		\textbf{ResNet50} & --- & 8 & 10 \\
		\textbf{MobileFaceNet} & 0 & --- & 10 \\
		\textbf{ShuffleFaceNet} & 0  & 0 & --- \\
		\bottomrule
	\end{tabular}
\end{table}

\paragraph{Prompt selection.}
We analyzed the influence of prompt formulation on retrieval performance using the mAP metric. As shown in Table \ref{tab:face_compare_by_promt}, Prompt 3 and Prompt 1 consistently outperform Prompt 2. While Prompt 1 and Prompt 3 achieve an equal number of total wins (8 out of 12), Prompt 3 holds a direct advantage over Prompt 1 (5 wins vs. 2). This suggests that the more detailed or structured phrasing in Prompt 3 provides slightly more discriminative textual representations for facial attributes, though the margin is narrower than in the tattoo scenarios.
\begin{table}[htbp]
	\centering
	\caption{Pairwise statistical comparison between prompts for face sketch retrieval. Each cell indicates the number of times (out of 12 comparisons) that the prompt in the row was significantly superior to the prompt in the column.}
	\label{tab:face_compare_by_promt}
	\begin{tabular}{cccc}
		\toprule
		& \textbf{Prompt 1} & \textbf{Prompt 2} & \textbf{Prompt 3} \\
		\midrule
		\textbf{Prompt 1} & --- & 6 & 2 \\
		\textbf{Prompt 2} & 4 & --- & 1 \\
		\textbf{Prompt 3} & 5 & 3 & --- \\
		\bottomrule
	\end{tabular}
\end{table}

\begin{figure}[htbp]
	\centering
	\includegraphics[width=0.8\linewidth]{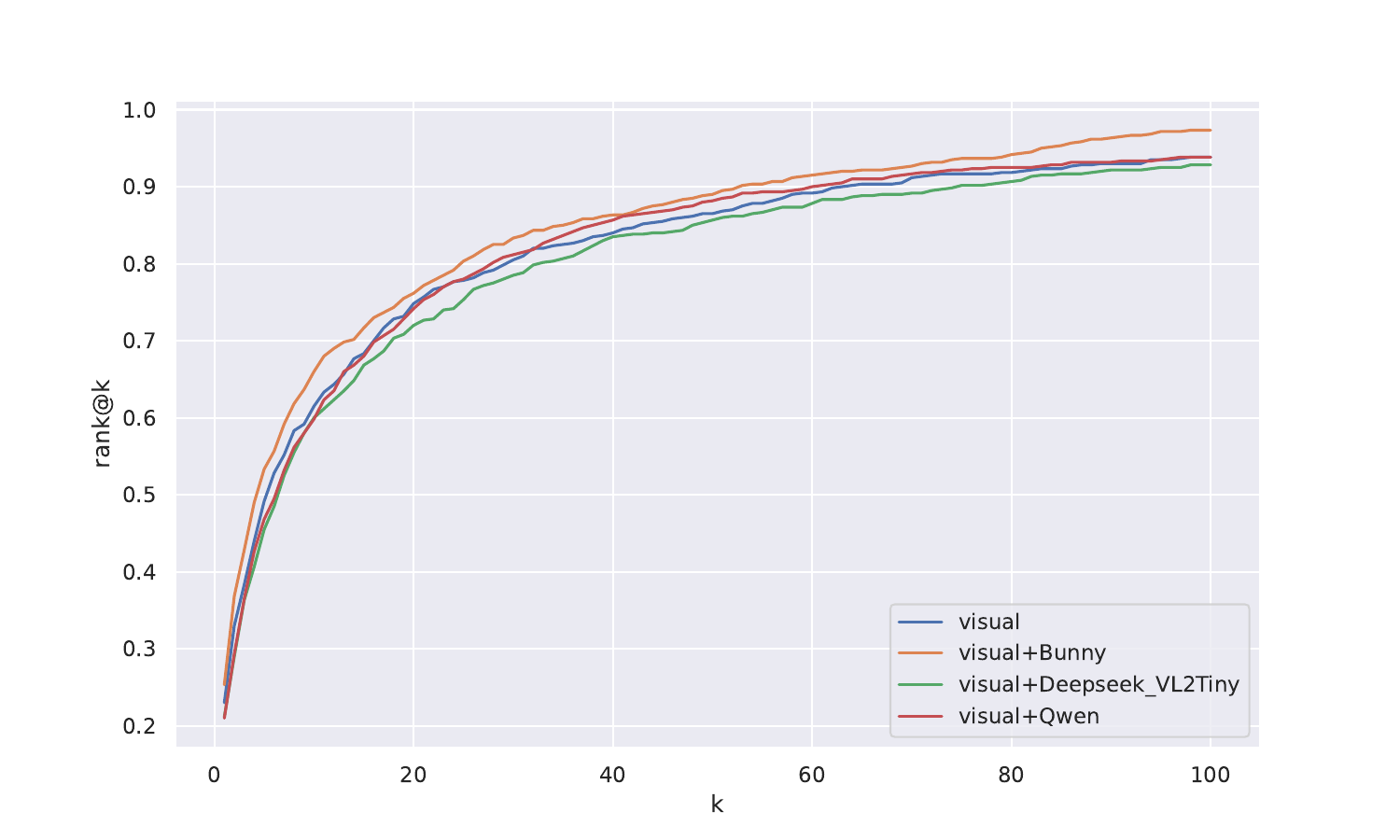}
	\caption{Comparison of standalone visual performance (ResNet50) and its combination with various MLLMs (Prompt 3) in terms of mAP.}
	\label{fig:visual_plus_llm_face}
\end{figure}

\paragraph{Impact of multimodal fusion.}
We then evaluated the benefit of augmenting the strong visual baseline (ResNet50) with MLLM-generated textual descriptions. Figure \ref{fig:visual_plus_llm_face} illustrates the mAP values for various MLLMs combined with the visual backbone. 

In contrast to the tattoo retrieval scenarios, where textual augmentation yielded widespread significant improvements, the face sketch scenario shows more limited gains. Table \ref{tab:face_stat_results} reports the performance of the top configurations. Only one combination (ResNet50\_Bunny-v1.1-4B\_Prompt3) achieves a statistically significant improvement over the visual-only baseline (mAP 0.353 vs. 0.384). The remaining combinations provide marginal or non-significant gains. This suggests that in face sketch retrieval, visual information remains the dominant source of discriminative power, and textual descriptions offer selective rather than universal benefits. This may be due to the high efficacy of existing face recognition backbones compared to the specialized tattoo domain.

\begin{table}[htbp]
	\centering
	\caption{Performance comparison of the best-performing methods for face sketch-to-photo retrieval. \ccheck indicates statistically significant improvement with respect to the visual feature alone.}
	\label{tab:face_stat_results}
	\begin{tabular}{lccccc}
		\toprule
		\textbf{Method} & \textbf{mAP} & \textbf{Rank@1} & \textbf{Rank@10} & \textbf{Rank@20} & \textbf{Rank@50} \\
		\midrule
		ResNet50 (Base) & 0.353 & 0.230 & 0.615 &  0.748 & 0.865 \\
		ResNet50\_Bunny-v1.1-4B\_Prompt3 & \textbf{0.384}\ccheck & \textbf{0.253} & \textbf{0.660}\ccheck & \textbf{0.762} & \textbf{0.890}\ccheck \\
		ResNet50\_DeepSeek-VL2-tiny\_Prompt1 & 0.364 & 0.233 & 0.625 & 0.730 & 0.873 \\
		ResNet50\_Bunny-v1.1-4B\_Prompt1 & 0.357 & 0.232 & 0.625 & 0.758 & 0.887 \\
		ResNet50\_Qwen2-VL-2B\_Prompt3 & 0.333 & 0.210 & 0.598 & 0.742 & 0.882 \\
		\bottomrule
	\end{tabular}
\end{table}

\paragraph{Optimal configurations and operational relevance.}
The best configuration (ResNet50\_Bunny-v1.1-4B\_Prompt3) achieves a Rank@1 of 0.253. While this represents an improvement over the visual baseline, it is notably lower than the Rank@1 scores observed in tattoo sketch retrieval (0.649). This discrepancy highlights the inherent difficulty of face sketch matching, where subtle stylistic variations in facial features can drastically alter similarity scores. 
Operationally, these results indicate that while the multimodal framework provides a measurable boost, it should not be relied upon as a standalone identification tool in face cases. The low Rank@1 suggests that investigators must treat the output as a prioritized shortlist rather than a definitive match. However, the significant improvement in Rank@10 (0.615 to 0.660) confirms that the textual component helps promote relevant candidates into the top tier of results, reducing the manual effort required to locate potential matches within a larger database.

	
\subsection{Discussion: Implications for forensic practice}

The experimental results validate the core premise of this work: a unified multimodal retrieval framework can effectively handle heterogeneous forensic evidence, ranging from high-quality photographs to ambiguous sketches and textual descriptions. While performance varies across scenarios, the consistent benefits of semantic enrichment and the flexibility of the architecture offer significant operational value. This section discusses the implications of these findings, addressing the performance disparities, the role of model efficiency, human factors, and ethical considerations.

\paragraph{Semantic diversity and methodological scope.}
The performance disparity between the tattoo scenarios (mAP up to 0.96) and the face sketch scenario (mAP $\approx$ 0.38) is primarily attributable to the inherent semantic richness of the biometric traits rather than solely the visual backbones. Tattoos exhibit high semantic diversity, encompassing distinct objects, symbols, text, and artistic styles that are readily captured and discriminated by textual descriptions. In contrast, facial structures are anatomically standardized; variations are often subtle and difficult to articulate discriminatively in natural language. This distinction is evident in the photo-to-photo tattoo results (Tables \ref{tab:photo_webtattoo_map} and \ref{tab:photo_bivtatt_map}), where generic backbones (MobileNetV2) and domain-specific ones (TattTRN) achieved comparable baseline performance and similar relative gains from textual fusion. This indicates that for tattoos, the semantic component itself is highly discriminative regardless of the visual encoder. Conversely, in face retrieval, the limited semantic variance reduces the utility of textual augmentation, as natural language struggles to capture the fine-grained geometric nuances required for facial identification. Consequently, the inclusion of the face sketch scenario serves a critical methodological purpose: it acts as a contrast to define the operational scope of the framework. By comparing a semantically rich domain (tattoos) against a semantically constrained one (faces), we establish that the proposed approach yields optimal results when applied to forensic tasks with wide and clear semantic variability. In contrast, for semantically narrow domains (e.g., ear biometrics or specific scar patterns), the benefits of language-mediated retrieval may be diminished. Thus, the face scenario delineates the boundaries of applicability, guiding practitioners toward deploying this framework where semantic diversity can be fully leveraged.

\paragraph{Limitations and real-world variability.}
While the sketch-to-photo results are promising, they were obtained using a controlled dataset where sketches were drawn by experts viewing the source image. In real forensic investigations, sketches are often generated from witness memory, introducing greater visual distortion and stylistic variability. This suggests that the visual performance reported here (particularly for TattTRN) may be optimistic compared to operational deployment. However, this limitation further underscores the value of the proposed multimodal approach: as visual fidelity decreases in real-world memory-based sketches, the semantic stability of textual descriptions becomes increasingly vital. 

\paragraph{Model efficiency and output verbosity.}
A consistent finding across all scenarios is that smaller, efficient MLLMs (specifically DeepSeek-VL2-Tiny (1B) and Bunny-v1.1-4B) outperformed larger models like Qwen2.5-VL-7B or LLaVA-1.6-7B. This counter-intuitive result aligns with 
previous findings indicating a strong negative correlation between model output verbosity and retrieval accuracy \cite{MLLMTattoo_CIARP2025}. Larger models tend to generate longer, more narrative descriptions that introduce noise into the embedding space, whereas smaller models produce concise, attribute-focused captions that align better with discriminative retrieval tasks. For forensic deployment, this suggests that computational efficiency and caption quality are more valuable than model scale, allowing for faster processing times and lower infrastructure costs without sacrificing accuracy.

\paragraph{Ethical considerations and reliability.}
Despite its potential, semantic retrieval introduces important ethical and operational challenges. Human descriptions are inherently subjective and may reflect perceptual bias, especially regarding soft biometric traits. Additionally, MLLM-generated captions may contain hallucinated or over-specified details, potentially introducing systematic errors into the indexing process. Increased descriptive detail may also amplify incorrect or biased information. Careful prompt design, validation procedures, and human oversight remain essential to ensure reliability and forensic accountability.

\paragraph{Human factors and training requirements.}
The human description-to-photo scenario revealed that variability in human input can outweigh algorithmic choices. The significant performance gap between expert annotators (Section \ref{sec:exp_human_description}) underscores that the effectiveness of semantic retrieval systems is contingent on the quality of the textual query. This highlights a critical training requirement for practitioners: personnel generating forensic descriptions must be guided by standardized protocols to minimize annotation bias and vocabulary inconsistency. Technology alone cannot compensate for ambiguous or inconsistent human input; therefore, operational deployment should pair this retrieval framework with structured guidelines for witness interviewing and evidence description.

Overall, the findings indicate that multimodal semantic retrieval can enhance forensic search capabilities, particularly in text-driven scenarios, while also underscoring the need for controlled integration, bias mitigation strategies, and practitioner training.


\section{Conclusion and future work}

This paper presented a multimodal retrieval framework designed to address the heterogeneity of forensic evidence. By integrating semantic descriptions generated by Multimodal Large Language Models (MLLMs) with domain-specific visual features, the proposed pipeline enables consistent retrieval across diverse scenarios, including photo-to-photo matching, sketch-to-photo identification, and human description-based search. The experimental evaluation demonstrates that transforming visual evidence into a unified textual space significantly enhances retrieval robustness, particularly when visual data is incomplete, degraded, or stylized.

The results highlight several key findings. First, in tattoo retrieval scenarios, the combination of textual and visual embeddings achieved state-of-the-art performance (mAP up to 0.96), validating the synergy between semantic captions and visual features in semantically rich domains. Second, the framework proved effective even in the absence of visual queries, with human description-to-photo retrieval achieving competitive accuracy (mAP $\approx$ 0.74), offering a viable tool for early-stage investigations. Third, the performance contrast with face sketch retrieval (mAP $\approx$ 0.38) delineates the methodological scope of the framework. While the architecture is flexible enough to handle multiple modalities, the disparity indicates that accuracy is contingent on semantic richness; tasks with high semantic diversity yield substantial gains, whereas semantically constrained domains offer limited benefits from textual augmentation. Finally, contrary to common assumptions, smaller MLLMs (DeepSeek-VL2-Tiny and Bunny-v1.1-4B) consistently outperformed larger counterparts, suggesting that concise, attribute-focused captions are more discriminative for retrieval than verbose narrative descriptions.

Despite these contributions, several limitations remain. The approach relies on the consistency and accuracy of automatically generated descriptions, which may vary across models and prompts. Additionally, the evaluation was conducted on controlled datasets; operational environments may introduce greater linguistic variability and noise in witness descriptions. Finally, the current study focuses on English-language prompts and a limited set of forensic markers.

Future work will address robustness to linguistic variability, cross-lingual retrieval, and adaptive prompt strategies. Further investigation is also required to assess system performance under real-world operational constraints, including incomplete, ambiguous, or partially incorrect descriptions. Expanding the methodology to additional forensic traits and exploring uncertainty-aware retrieval mechanisms constitute promising research directions.

Overall, the findings suggest that semantic-multimodal retrieval is a viable complementary paradigm for forensic search, with clear benefits in text-driven scenarios and defined boundaries in visually dominated tasks.

\section*{Declaration of generative AI and AI-assisted technologies in the manuscript preparation process}
During the preparation of this work the authors used Qwen3.5 via the Alibaba Cloud platform in order to improve the English language and clarify technical explanations, as none of the authors are native English speakers. After using this tool/service, the authors reviewed and edited the content as needed and take full responsibility for the content of the published article.


\printcredits

\bibliographystyle{cas-model2-names}

\bibliography{cas-refs}


%
%
%

\end{document}